\documentclass[
]{ceurart}

\usepackage{listings}
\begin{document}

\copyrightyear{2023}
\copyrightclause{
Copyright for this paper by its authors.
Use permitted under Creative Commons License Attribution 4.0 International (CC BY 4.0).
}

\conference{
IntRS'23: Joint Workshop on Interfaces and Human Decision Making for Recommender Systems, September 18, 2023, Singapore (hybrid event)
}

\title{GPT as a Baseline for Recommendation Explanation Texts}

\author[1]{Joyce Zhou}[%
orcid=0000-0003-1205-3970,
email=jz549@cornell.edu,
url=https://cephcyn.github.io/,
]
\cormark[1]
\author[1]{Thorsten Joachims}[%
orcid=0000-0003-3654-3683,
email=tj36@cornell.edu,
url=https://www.joachims.org,
]
\address[1]{Cornell University}

\cortext[1]{Corresponding author.}

\begin{abstract}
In this work, we establish a baseline potential for how modern model-generated text explanations of movie recommendations may help users, and explore what different components of these text explanations that users like or dislike, especially in contrast to existing human movie reviews.
We found that participants gave no significantly different rankings between movies, nor did they give significantly different individual quality scores to reviews of movies that they had never seen before.
However, participants did mark reviews as significantly better when they were movies they had seen before.
We also explore specific aspects of movie review texts that participants marked as important for each quality.
Overall, we establish that modern LLMs are a promising source of recommendation explanations, and we intend on further exploring personalizable text explanations in the future.
\end{abstract}


\maketitle

\section{Introduction}

ChatGPT and other generative large language models are becoming prominently capable of creating comprehensible and seemingly meaningful texts \cite{brownLanguageModelsAre2020, yangHarnessingPowerLLMs2023}, as well as being increasingly popular for casual users to interact with.
Furthermore, explainable AI (and explainable recommendation systems) have been an increasingly common topic of interest, with a wide range of goals including user satisfaction, personalization, and transparency \cite{tintarevSurveyExplanationsRecommender2007, tintarevEffectiveExplanationsRecommendations2007, tintarevEvaluatingEffectivenessExplanations2012, zhangExplainableRecommendationSurvey2020, chenMeasuringWhyRecommender2022}.
However, the potential for LLM-generated text explanations to augment recommendation system interfaces has not been explored with modern models much yet.

In this work, we establish a baseline potential for how modern model-generated text explanations of movie recommendations may help users, and explore what different components of these text explanations that users like or dislike, especially in contrast to existing human-written movie reviews.
We do this by surveying 120 participants and asking them to rank their movie preferences based on LLM-generated v.s. human-written movie review texts in a mockup recommendation environment, as well as give individual feedback for how accurate, informative, persuasive, and interesting each review text is.
We found that participants gave no significantly different rankings between movies, nor did they give significantly different individual quality scores to reviews of movies that they had never seen before.
However, participants did mark model-generated review texts as significantly better than human-written reviews when they were movies they had seen before.
We also explore specific aspects of movie review texts that participants marked as important or crippling for each quality.
These include structured item features such as genre or famous actors, but they also include particular interests such as special effects or historical relevance, cultural context of a work, or personal experiences of the review-writer themselves.

Overall, we establish that modern LLMs are a promising source of post-hoc explanations that could accompany item recommendations with relevant summaries to improve user satisfaction.
We intend on further exploring user-personalized text explanations in the future.

\section{Related Work}

There exists a substantial amount of past work in explanations for recommender systems \cite{vultureanu-albisiSurveyEffectsAdding2022}, which historically mostly used feature-based explanations \cite{tintarevSurveyExplanationsRecommender2007}.
More recently, there has been work on sentence generation or longer-form text generation for recommender system explanations.
Some of this work makes use of crowdsourced texts and extracts only relevant segments to show together with a recommendation \cite{niJustifyingRecommendationsUsing2019}, or offers personalized messages based on how similar another user's preferences are \cite{liPersonalizedPromptLearning2023}.
Other recommender system designs have been proposed that incorporate text explanations as part of the recommendation task itself, to make the recommender system more easily explainable from the start.
For example, a recommender system may be trained to predict review scores for some user-item pair, based on past user and item reviews \cite{zhengJointDeepModeling2017}, extract specific product features \cite{zhangExplicitFactorModels2014}, or predict the exact text that a user writes for a specific recommendation \cite{costaAutomaticGenerationNatural2018, hadaReXPlugExplainableRecommendation2021}.
However, most of this work has measured success through recommendation performance and general accuracy, not user experience.

There has also been substantial work with conversational recommender systems \cite{gaoAdvancesChallengesConversational2021}, which arguably incorporate some kind of explanation in the form of the conversation itself.
For example, \cite{zhaoPersonalizedReasonGeneration2019} build a full song recommendation chatbot, which serves songs to users together with a chat message sampled from a dataset of previously written human messages, evaluated on text coherence and click-through rate.

Most efforts with personalizing a text explanation assume that the text should be personalized to contain features that a reader strongly cares about, and not so much the text format itself.
Work on personalized explanations often evaluates for text quality and recommendation success rate, but often also includes user satisfaction.
\cite{liPersonalizedPromptLearning2023} build a system that generates personalized natural language explanations for a given user-item recommendation, and evaluate on text quality and explanation quality (based on item features).
\cite{changCrowdBasedPersonalizedNatural2016} uses crowdsourced reviews to generate (sample from human texts) personalized explanations for recommendations.

\section{Survey Design}


\subsection{Research Questions}

We started with the following research questions:

\begin{enumerate}{
    \renewcommand\labelenumi{\textbf{RQ\theenumi:}}
    
    \item \label{h:rankpref} How does presenting a movie (especially those that participants have not seen before) with model-generated texts, in contrast to human-generated texts, impact their watch preference rankings if at all?
    
    \item \label{h:likert} Do model-generated texts receive individual quality scores (for whether they are informative, persuasive, and interesting) on par with those of human-generated texts?
}\end{enumerate}

We highlighted rankings in unseen movies specifically because participants who have already seen a movie are likely to have more concrete opinions on whether they like it, which are unlikely to be affected by reading one new review text.
In contrast, individual quality scores are intended to focus on the text contents alone.


\subsection{Dataset}

To show a set of mildly customizable movie suggestions and texts for each participant, we needed to collect a broad set of relatively well-known movies to base suggestions on (the "seed" movie), a set of suggestions for each seed movie, and a set of human-generated and model-generated texts for each suggested movie.

We manually selected 10 well-known seed movies to ideally cover a range of movie genres, as well as raise the chance of most participants recognizing at least one seed movie.
For each seed movie, we collected a sorted list of similar movies using the first 5 pages of "similar movies" on MovieLens followed by 1 page of "similar movies" on BestSimilar.
For each movie suggestion, we collected 5 human-generated reviews to use as a baseline comparison.
These reviews consisted of the top 5 most "featured" non-spoiler reviews on IMDB\footnote{as of 2023/04/26, "featured" reviews seem to use a metric combining high helpfulness vote percentage and high total vote counts} for each movie.
For each movie suggestion, we also synthesized one model-generated text using the 5 human-generated reviews as a source of information about the movie.
Finally, we took the human-generated and model-generated texts, reformatted them to remove all linebreaks and extraneous whitespace, and truncated them to 100 tokens when necessary (with "..." append in these cases to indicate that further text was not shown).

In summary, there are 10 well-known seed movies, 5 lesser-known suggestions based on each seed movie (50 total), 5 human-generated reviews for each suggestion (250 total), and 1 model-generated review for each suggestion (50 total).

\subsection{Survey Procedure}

We generated condition codes to ensure balanced representation between human-generated and bot-generated texts for each movie, as well as between different human authors of the human-generated texts.
Each participant was assigned one condition code that determines which of the 6 review texts is shown for each individual movie suggestion, as well as if they are shown sources for all review texts.

Survey participants were asked to select one seed movie, then shown 5 suggested movies together with texts describing each suggestion.
They were asked to rank suggestions, as well as mark which suggested movies they had seen before.
Finally, for each suggestion, they were asked to rate how accurate (only if they had seen the movie before), informative, persuasive, and interesting they found the texts, as well as elaborate on what aspects of the text satisfied or did not satisfy these attributes.

We recruited 120 participants from Amazon Mechanical Turk, limiting to only workers who are within the United States, have prior task approval rating of at least 95\%, and have a minimum of 1000 approved tasks.
Study participants received a minimum of \$2.00 for completing the survey, with an additional bonus up to \$5.50 based on time spent within the survey to ideally reach a \$15.00 hourly wage.

Further fine detail about survey setup and procedure is available on our dataset repository\footnote{https://github.com/cephcyn/gpt-reccexpl-mturksurvey}.

\section{Results}

Overall, we found no significant difference between movie preference rankings with model-generated texts in contrast to human-generated texts (RQ\ref{h:rankpref}), even when focusing on rankings of solely unseen movies.
We also found no significant difference overall in Likert quality scores between model-generated and human-generated review texts (RQ\ref{h:likert}) for movies that participants had never seen before. 
However, Likert quality scores of model-generated texts for movies that participants \textit{had seen} before were significantly higher than those of human-generated texts.

\subsection{Preference Rankings}

We saw no significant effect of showing model-generated or human-generated texts on movie preference rankings (related to RQ\ref{h:rankpref}).

In Table \ref{tab:ranking_top}, we show the number of top-ranked movies distributed across different text sources.
To remove effects of some movies already being seen, we also show this comparison for the "unseen only" rankings (removing all seen movies, as well as removing rankings of respondents who had 4 or more seen movies) and "seen only" rankings (same criteria but removing unseen movies).
Ranking sets which had too many seen or unseen movies to have any candidate for "top-ranked movie" were excluded.

We were also interested in potential differences in average ranking position.
As the "unseen only" and "seen only" filters often remove at least one movie from the ranking, we chose to measure average ranking by first applying any filters, then normalizing all remaining movies such that they are ranked with decimal values from 0 (top-ranked) to 1 (lowest-ranked).
In Table \ref{tab:ranking_avgnorm}, we show an average ranking for all movies shown with different text sources, across all respondents.
Note that across all three "un/seen" condition filters, movies with model-generated texts are ranked marginally higher on average, but none of these differences are significant.


\begin{table}
    \caption{Number of top-rank choices for movies assigned model vs. human texts (n=120)}
    \label{tab:ranking_top}
    \begin{tabular}{c || c | c | c | c}
        \toprule
        Filter & Human & Model & Excluded & P-value \\
        \midrule
        All         & 57 & 63 & -  & 0.64 \\ 
        Unseen only & 57 & 53 & 10 & 0.77 \\
        Seen only   & 12 & 17 & 91 & 0.45 \\
        \bottomrule
    \end{tabular}
\end{table}

\begin{table}
    \caption{Average ranking (normalized; 0 is top-ranked, 1 is lowest-ranked) for movies assigned model vs. human texts}
    \label{tab:ranking_avgnorm}
    \begin{tabular}{c || c | c | c}
        \toprule
        Filter & Human & Model & P-value \\
        \midrule
        All         & 0.513 (n=300) & 0.486 (n=300) & 0.35 \\ 
        Unseen only & 0.515 (n=220) & 0.488 (n=196) & 0.47 \\
        Seen only   & 0.544 (n=45)  & 0.449 (n=50)  & 0.26 \\
        \bottomrule
    \end{tabular}
\end{table}

In summary, While model-generated texts did not encourage participants to rank movies higher, they were not ranked significantly lower either!

As a side note, this is a minor demonstration that it is possible to use model-generated texts in review style without significantly altering media preference rankings.
Admittedly, we gave GPT a specific prompt to write a friendly review without the intention of promoting any particular item.
A system designer with different goals would still be entirely able to prompt GPT to heavily promote or heavily criticize a particular piece of media.

\subsection{Likert Quality Scores}

Because a movie ranking position is not necessarily tied to how relevant each shown text is, we also asked participants to give individual feedback for each text\footnote{We found no correlation between Likert quality scores and movie ranking position for all qualities other than "persuasiveness", and then only marginal.}.
We saw no significant differences between three main text qualities (informative, persuasive, and interesting) for model-generated and human-generated texts across all movies (RQ\ref{h:likert}).

\begin{table}
    \caption{Average Likert score (5 is agree, 1 is disagree) for movies assigned model vs. human texts. P-values were calculated with a logistic regression model, treating Likert scores as ordinal.}
    \label{tab:likert_pscores}
    \begin{tabular}{c | c || c | c | c}
        \toprule
        Quality & Filter & Human & Model & P-value (binomial) \\
        \midrule
        Accurate      & Seen only   & 4.063 (n=79)  & 4.284 (n=102) & 0.02 \\ 
        \midrule
        Informational & All         & 3.943 (n=300) & 4.053 (n=300) & 0.49 \\ 
        Informational & Unseen only & 3.972 (n=221) & 3.984 (n=198) & 0.53 \\
        Informational & Seen only   & 3.860 (n=79)  & 4.186 (n=102) & 0.03 \\
        \midrule
        Persuasive    & All         & 3.850 (n=300) & 3.966 (n=300) & 0.40 \\ 
        Persuasive    & Unseen only & 3.877 (n=221) & 3.863 (n=198) & 0.35 \\
        Persuasive    & Seen only   & 3.772 (n=79)  & 4.166 (n=102) & 0.002 \\
        \midrule
        Interesting   & All         & 3.913 (n=300) & 4.050 (n=300) & 0.10 \\ 
        Interesting   & Unseen only & 3.877 (n=221) & 3.904 (n=198) & 0.86 \\
        Interesting   & Seen only   & 4.012 (n=79)  & 4.333 (n=102) & 0.003 \\
        \bottomrule
    \end{tabular}
\end{table}


\begin{figure}
    \centering
    \includegraphics[width=\linewidth]{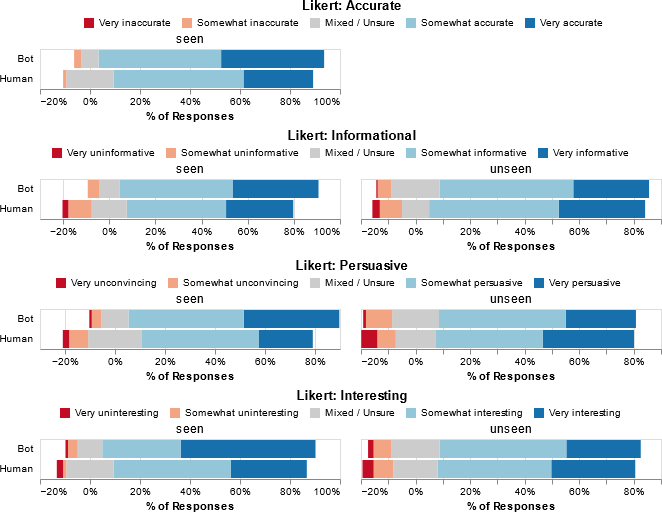}
    \caption{
        Quality rating distribution across all movie responses, split by seen status.
        Left column is responses for already-seen films only, right side is responses for unseen films only. 
        Note that there is no distribution for accuracy responses on unseen films, as the accuracy rating question was only shown for texts describing films that respondents marked as seen.
    }
    \label{fig:likert_filtered}
\end{figure}

However, when filtering solely on \textit{seen} movie responses, we found that model-generated texts received significantly higher Likert quality scores than human-generated texts (Figure \ref{fig:likert_filtered}) (Table \ref{tab:likert_pscores}).
This is an especially odd effect when we contrast marginal differences between the distribution of responses for seen vs. unseen movies.
For seen films, model-generated text seems to get more strongly-decided rankings, while in unseen films it is human-generated text that is ranked more strongly instead.


Again, we generally observe that model-generated texts are ranked at least as well as human-generated texts across a range of criteria, even without any level of customization for each participant.

\subsection{Qualititative Analysis of Quality Responses}

To explore what aspects of each review participants may particularly like or dislike, as well as how these aspects may relate to different qualities that make a review text appealing, we read through the qualitative responses that participants gave and summarize them here.

In general, participants mentioned several recurring topics, regardless of which attribute we were asking about.
These common topics are:
\begin{itemize}
    \item \textit{Raw release information and movie synopsis}.
    This includes plot summary (often avoiding spoilers), story themes, content warnings, genre, setting, director or actor names, or technical details such as special effects or format.
    Participants tended to criticize texts missing this information: "It didn't really tell me what the movie was about. The writer just said how much they like the movie."
    
    \item \textit{Context around movie production or release}.
    This includes descriptions of critical reception, awards, contrasts against other period-specific or genre-overlapping films, commentary on how other reviewers are correct or incorrect, historical impact, or how a movie otherwise fits into a larger body of work.
    
    \item \textit{Opinion and personal movie experience}.
    This includes personal opinions, emotional responses that reviewers had, or otherwise general praise or criticisms of a movie.
    For instance, participants highlighted how texts describe the "strong performance" of an actor, how a film is "fascinating and compelling", or how a review text "contained actual criticism".
    
    \item \textit{Review structure and writing style}.
    Several participants commented on the overall structure of a review text.
    This was usually criticism about texts focusing solely on a reviewer's personal experiences or anecdotes instead of information about the movie itself.
    Participants often also described review writing style: whether it was short or wordy, the tone or emotion of a review, or what kind of colorful language was used to describe a movie.
\end{itemize}

Across these responses, we noticed that human-generated review texts sometimes focused on personal experiences (occasionally excluding a plot summary entirely), while model-generated texts usually included a plot summary (described by respondents as vague or spoiler-free) and summary of critical reception to a movie.

Participants tended to emphasize different attribute types more in different question types.
With accuracy, participants focused on raw release information and release context more than they did on subjective experiences, except for when a participant specifically said they agreed with that experience.
With informativeness, participants often summarized review attributes across all categories, and sometimes criticized reviews for saying nothing objective about a movie or being overly vague.
With persuasiveness, there was more variation across participant responses.
Some highlighted how a review conveys enthusiasm about a movie (or other personal experiences) or appreciated a review framing the achievements of a movie against others from the same era, while other participants criticized a review for containing too much subjective information and not enough plot summary.
Finally, with interestingness, participants tended to focus more often on writing style or creative wording overall, as well as criticizing vagueness.
However, like persuasiveness, there was a good amount of variation between participants for what other topics they found particularly interesting.

\section{Discussion \& Future Directions}

Overall, we established a baseline of model-generated recommendation texts being possible and promising!
In addition, a good number of the survey participants responded to the final question saying they did want to see more reviews from the same authors, even for the ones that were model-generated.

The Likert scale effect being stronger for movies that were seen already was unexpected and interesting.
However, in retrospect, this makes sense for accuracy.
Multiple respondents mentioned how a human-generated text contained only anecdotes or personal experiences with a movie and nothing that can be validated.
For the other qualities, we suspect this is due to similar reasons: model-generated texts tend to summarize general opinions and offer vague summaries of a movie plot and critical reception, without sharing any strong personal opinion.


One big weakness of this work is that GPT texts were still generated using human reviews.
This is both a weakness for movies that may be recommended that have no human reviews written yet, as well as a general ethical flaw of LLMs (as they are often built atop unpaid human labor with model training and model input).
We do not demonstrate any capacity to write competent reviews for media that have no already-written human reviews here, and this is a future work direction that deserves attention.

Finally, one major future work direction that we are especially interested in includes exploring how we can allow users to customize different review attributes or otherwise learn these preferences to better generate summary texts that improve user satisfaction and focus on what they prioritize.

\section{Conclusion}

Existing recommender systems have tried incorporating chatbots and customized text explanations, but have not had landmark well-performing LLMs to make use of until now.
We ran a survey that simulates a very basic recommender system, and a barely-customized LLM, but establishes that LLM-generated text explanations could be a good substitute, supplement, or complement to existing human review texts.
We found that while LLM review text is not performing significantly better in rankings or direct scoring compared to human texts for unseen movies, it is also not doing any worse.
Ultimately, this is a good baseline to build on with future work.
Based on the freeform responses we got from survey participants, there is a good amount of personalization that could be provided to individual reccommender system users that an explanation-text-offering LLM could provide.

\begin{acknowledgments}
This research was supported in part by the Graduate Fellowships for STEM Diversity (GFSD), as well as NSF Awards IIS-1901168 and IIS-2008139.
All content represents the opinion of the authors, which is not necessarily shared or endorsed by their respective employers and/or sponsors.

\end{acknowledgments}

\bibliography{ref}




\end{document}